\pdfoutput=1

\documentclass[11pt]{article}

\usepackage[preprint]{acl}

\usepackage{times}
\usepackage{latexsym}

\usepackage[T1]{fontenc}

\usepackage[utf8]{inputenc}
\usepackage{xcolor} 
\usepackage{soul} 
\usepackage{enumitem}
\usepackage{array}        
\usepackage{makecell}     
\usepackage{microtype}

\usepackage{inconsolata}

\usepackage{graphicx}
\usepackage{multirow}
\usepackage{amsmath}
\usepackage{booktabs}
\usepackage{arydshln}
\usepackage{subcaption}

\title{Assessing the Sensitivity and Alignment of FOL Closeness Metrics}

\author{
 \textbf{Ramya Keerthy Thatikonda\textsuperscript{$\forall$}} \ \ \
 \textbf{Wray Buntine\textsuperscript{$\forall$,$\exists$}} \ \ \
 \textbf{Ehsan Shareghi\textsuperscript{$\forall$}}
\\
\\
 \textsuperscript{$\forall$}Department of Data Science \& AI, Monash University
 \\
 \textsuperscript{$\exists$}College of Engineering and Computer Science, VinUniversity
\\
}

\begin{document}
\maketitle
\begin{abstract}
The recent successful paradigm of solving logical reasoning problems with tool-augmented large language models (LLMs) leverages translation of natural language (NL) statements into First-Order Logic~(FOL) and external theorem provers. However, the correctness of FOL statements, comprising operators and text, often go unverified due to the lack of a reliable evaluation metric for comparing generated and ground-truth FOLs.
In this paper, we conduct a comprehensive study on the \emph{sensitivity} of existing NL-, FOL-, and graph-based metrics to capture differences between a sampled FOL and its corresponding ground-truth. We then measure the \emph{alignment} between a metric-based ranking of FOL outputs and a strong LLM as-a-judge. To do this, we first apply operator and text-based perturbations to ground-truth FOL statements to assess metric sensitivity. We then evaluate metric robustness by comparing the metrics against LLMs judgment.
Our empirical findings highlight a clear oversensitivity in the n-gram metric BLEU for text perturbations. The operator perturbation affects the semantic graph metric Smatch++ for structural changes, and the FOL metric for specific operator changes. We observe a closer alignment between BertScore and LLM judgement, proving the importance of semantic evaluation.
Additionally, we  show that combining metrics enhances both robustness and sensitivity compared to using individual metrics. \footnote{Our code is available at \url{https://github.com/RamyaKeerthy/AlignmentFOL}}
\end{abstract}

\section{Introduction}
Large language models (LLMs) have advanced natural language reasoning, but logical and mathematical reasoning have long relied on formal, structured languages for proving deductions and theorems, a process that predates deep neural networks \cite{quinonero2006machine}. This approach remains relevant today, especially for reasoning tasks that can be solved using formal statements. In case of first-order logic (FOL), LLM generations are used as intermediate steps and subsequently passed to theorem provers to solve the problem~\cite{logiclm, satlm, linc}. 
Compared to the Chain-of-Thought~(CoT) approach \cite{chainofthought}, where the model first reasons and then solves, FOL generation demonstrated superior reliability by offloading the reasoning task to an external tool. Translating natural language (NL) into FOL enhanced the overall rigor of the process.

Generating FOL from NL is a challenging task that tests the ability of LLMs to accurately interpret and convert informal language into a formal, structured token sequence ~\cite{yang2024formal,wu2022autoformalization}. The lack of ground truth for FOL generations complicates direct verification. \citet{logicllama} addressed this challenge by developing a system specifically for FOL generation, incorporating an operator-based evaluator to rate the outputs. This evaluation is combined with BLEU score, using a threshold as a metric in a reward model. However, the reliance on thresholds complicates the interpretation of translation quality. Manually assessing formal logic generations is labor-intensive and has received relatively less attention compared to traditional text translation metrics. 

In this work, we analyze existing natural language translation, tree, and graph evaluation metrics, focusing on those that offer strong sentence-level comparisons. We establish a framework to systematically introduce perturbations and analyze the existing metrics in the presence of these anomalies in formal language, specifically first-order logic. Metric sensitivity helps reveal the degree to which each metric responds to specific types of perturbations. 
To further assess these metrics on real-world examples, we generate sample FOLs for NL statements in FOLIO~\cite{folio} and MALLS~\cite{logicllama} datasets using an LLM and rank them against ground truth values. The ranking is conducted using established metrics, and LLM-based evaluators. After a small-scale experiment to select the LLM most aligned with human judgment for this task, we measure how well each metric-based ranking aligns with the selected LLM’s judgments \footnote{While using an LLM as a judge is one alternative, our alignment-based method offers a more resource-efficient approach by reducing the need to call an additional LLM.}. Our experiments reveal that BertScore has a better alignment as an individual metric, and its combination with other metrics could further boost its alignment. Our findings provide insights into the sensitivity of current metrics and their applicability to symbolic generation tasks.

\section{FOL Closeness Metrics}

Evaluation scores in natural language generation, such as BLEU \cite{bleu} and ROUGE \cite{rouge}, perform n-gram matching between reference text and candidate outputs. METEOR~\cite{meteor}, while also based on n-gram overlap, incorporates additional factors known to result in improved correlation with human judgments. BERTScore \cite{bertscore} leverages contextual embeddings generated by a pre-trained BERT~\cite{DBLP:conf/naacl/DevlinCLT19} to compute cosine similarity between sentences. 

In contrast, logical equivalence \cite{logicllama} evaluates FOL translations by comparing the truth values of formal statements, abstracting away from their textual semantics. Another relevant domain is Abstract Meaning Representation (AMR) graph metrics, which compare the structural similarity of semantic graphs. Given the structured nature of FOL statements, we leverage Smatch++ \cite{opitz2023smatch++}, which incorporates preprocessing, alignment, and sub-graph scoring. These metrics capture different dimensions of divergence between ground truth and translations: traditional metrics focus on surface-level and semantic discrepancies, while formal evaluation methods assess deeper logical consistency. We present results demonstrating how a representative set of these metrics respond to variations in logical constructs within formal language translations.

\section{Evaluation Framework}
\label{sec:perturbation}
\subsection{Perturbation Evaluations}
The effect of perturbations measures the \emph{sensitivity} of the metrics by assessing how small changes or variations in the FOL statements impact the metric scores. Based on the ground-truth of the FOLIO dataset~\cite{folio}, we utilize nine operators to construct a formal logic framework. 

To evaluate the performance of these metrics, we first conduct a self-matching experiment on the statements and normalize the results based on the variations observed in this process.
The perturbation strategies are chosen for sensitivity measure to answer questions drafted to provide insight into metric sensitivity. There are two variations, where the sensitivity can be measured by perturbing the operators or the text (predicates and variables):

\paragraph{Operator Quantifier:} \textit{How well does the metric respond when a universal quantifier is misinterpreted as existential?} In this perturbation, we swap the quantifiers $\forall$ and $\exists$ where applicable. For example, the formula $\forall\text{x} (\text{W}(\text{x}, \text{C}) \rightarrow \text{A}(\text{x}, \text{C}))$ becomes $\exists\text{x} (\text{W}(\text{x}, \text{C}) \rightarrow \text{A}(\text{x}, \text{C}))$. 
\vspace{-4pt} 
\paragraph{Operator Negation:} \textit{Will the metric be affected by the changes in negation logic?} 
In this perturbation, we either remove the negation of predicates, if present, or add it when absent. For example, $\forall\text{x} (\neg\text{W}(\text{x}, \text{C}) \rightarrow \text{A}(\text{x}, \text{C}))$ changes to $\forall\text{x} (\text{W}(\text{x}, \text{C}) \rightarrow \neg\text{A}(\text{x}, \text{C}))$.
\vspace{-4pt} 
\paragraph{Operator And/Or:} \textit{How well can the metric distinguish between conjunction and disjunction?} This perturbation involves a swap of logical operators, such as (And, Or). 
\vspace{-4pt}   
\paragraph{Text minus Operator:} \textit{To what extent does a metric rely on textual content?}
This perturbation focuses on the role of text without operators in influencing similarity scores. All logical operators are removed, and any multiple predicates are connected by a disjunction ($\lor$) to preserve the structure. For instance, $\forall\text{x} (\neg\text{W}(\text{x}, \text{C}) \rightarrow \text{A}(\text{x}, \text{C}))$ becomes $\text{W}(\text{x}, \text{C}) \lor \text{A}(\text{x}, \text{C})$.\footnote{
These errors pass through the tool without triggering any issues, making them a common occurrence in FOL generations by LLMs (Appendix~\ref{app:rationale}). Identifying this problem highlights a significant gap in the reliability of LLM-translated FOLs.}
    \vspace{-4pt} 
\paragraph{Text minus Variable:} \textit{How reliant are metrics on textual predicates over variable structure?}
All text values are replaced with generic variables and compared with the ground truth. For example, $\forall\text{x} (\neg\text{WantToBeAddictedTo}(\text{x}, \text{caffeine}) \rightarrow \text{AwareThatDrug}(\text{x}, \text{caffeine}))$ becomes $\forall\text{x} (\neg\text{A}(\text{x}, \text{C}) \rightarrow \text{B}(\text{x}, \text{C}))$. 
\footnote{All previous examples, except for text minus `Variable', have been shortened for space.}

\subsection{Sample Evaluations} 
Measuring the sample correctness with respect to the ground truth allows for an assessment of \emph{alignment} between different types of rankers. We extracted FOL statements from the FOLIO and MALLS dataset and implemented a sampling process in which \texttt{gpt-4o}~\cite{achiam2023gpt} was prompted in a zero-shot setting to generate three FOL samples for each natural language input. Each data point, consisting of a natural language statement and its corresponding FOL label $\{NL, FOL\}$, was used as input to \texttt{gpt-4o} (see Appendix~\ref{app:prompt} for prompt detail). The model then generated three candidate FOL statements: $\{FOL_1, FOL_2, FOL_3\}$. The generations with duplicate samples were removed, resulting in a total of 728 and 402 unique FOL outputs for FOLIO and MALLS respectively.

These samples were evaluated using various metrics by assigning scores to each comparison. In cases where two or more samples received identical scores, their ranks were adjusted accordingly. For example, if the FOLs were initially ranked [1, 2, 3], but $FOL_1$, and $FOL_2$ had equal scores, the ranks were updated to [1, 1, 3]. 

To evaluate the effectiveness of the metrics in ranking, we further used an LLM-based judge to assess the quality of the FOL samples, providing a broader perspective on the comparative rankings~(see Appendix~\ref{app:ranking} for prompt details).

\section{Experimental Setup}
\label{sec:exp}
\subsection{Data Preparation}
We use the training set of the FOLIO dataset (consisting 1001 records) for our experiments because of the availability of ground truth FOL in the dataset. Since our focus is on individual FOL statements, we decompose the records into single data points. We extract a total of 1689 records, ensuring a diverse combination of operators. To ensure the reliability of our results, we add MALLS dataset (sampled 1000 records) with individual FOL statements. 
The detailed data statistics are provided in Appendix~\ref{app:stats}.

\subsection{Evaluation Preparation}
\paragraph{Perturbation.}
The perturbations are evaluated using six metrics: BLEU (BL), ROUGE (RO), METEOR (ME), Logical Equivalence (LE), BERTScore (BS), and Smatch++ (SP). Following the method outlined by \citet{logicllama}, we first convert the FOL statements into a parsable format for each metric. For LE, an additional syntax check is conducted to ensure if the truth value of the FOL statement is valid before comparison. 
Due to the nature of the perturbations, they are applied only to relevant records. 
For example, quantifier perturbation is possible only if the statement contains a quantifier. The percentage of data affected by the perturbation is provided in Table \ref{tab:pertperc}.

\begin{table}\footnotesize
  \centering
        \resizebox{1\columnwidth}{!}{%
  \begin{tabular}{lcccccccc}
    \toprule
    & \rotatebox{90}{\textbf{op-Quant $\downarrow$}} 
    & \rotatebox{90}{\textbf{op-Neg $\downarrow$}} 
    & \rotatebox{90}{\textbf{op-AndOr $\downarrow$}} 
    & \rotatebox{90}{\textbf{t-Operator $\downarrow$}} 
    & \rotatebox{90}{\textbf{t-Variable $\downarrow$}} \\
    \hline
    \textbf{FOLIO}  & 61.40 & 99.88 & 54.29 & 99.00 & 99.70 \\
    \textbf{MALLS}  & 99.30 & 100.00 & 91.50 & 100.00 &100.00 \\
    \hline
  \end{tabular}}
  \caption{Percentage of perturbations applied to all FOL records. $\downarrow$ indicates preference for lower values.}
  \label{tab:pertperc}
  \vspace{-3mm}
\end{table}

\paragraph{LLM-generated FOL Samples.}
We use \texttt{gpt-4o} with temperature 0.6 and n=3 to generate three samples for each input (Appendix~\ref{app:prompt}). Samples that had two or more identical FOL statements were discarded, reducing the dataset to 728 and 402 records for FOLIO and MALLS respectively. 

\paragraph{LLM Selection for Judge.} For choosing an LLM to judge the alignment of these samples, we perform a small scale evaluation between 3 LLMs (gpt-4o, o3-mini, gemini-flash-2.0) as independent judges and 3 human annotators. Due to the cost of human annotations, we selected 87 records from FOLIO, covering all combination of operators. The experiments indicated a stronger alignment of o3-mini with human annotators (see Appendix~\ref{sec:annotator}). Hence we use o3-mini as the LLM judge in the following experiments.

\paragraph{Alignment Measure.} We use Root Mean Square Error (RMSE) to evaluate the alignment between two preferences (i.e.,  metric-based ranking vs. LLM-based ranking).  A \underline{lower} RMSE score indicates a better alignment. 

\section{Results and Discussion}
We present results from the two variations.
\paragraph{Perturbation Analysis.}
\begin{table}\footnotesize
  \centering
      \resizebox{1\columnwidth}{!}{%
  \begin{tabular}{p{0.2cm}p{1.8cm}p{0.45cm}p{0.45cm}p{0.45cm}p{0.45cm}p{0.45cm}p{0.45cm}}
    \toprule
    &\textbf{Pertb} & \textbf{BL}& \textbf{LE} & \textbf{RO}& \textbf{ME}& \textbf{BS} & \textbf{SP} \\
   \hline
    \multirow{5}{*}{\rotatebox{90}{FOLIO}}&op-Quanti $\downarrow$     & 0.96 & 0.97 & 0.96 & 0.96  & 0.99    & \textbf{0.95}      \\
    &op-negation $\downarrow$   & 0.68 & 0.77 & 0.93 & 0.83  & 0.96   & \textbf{0.38}    \\ 
    &op-AndOr $\downarrow$  & 0.88 & \textbf{0.74} & 0.96 & 0.96  & 0.99        & 0.93  \\
    &t-operator $\downarrow$  & \textbf{0.18} & 0.62 & 0.56 & 0.47  & 0.88        & 0.51 \\
    &t-variable $\downarrow$  & \textbf{0.25} & 0.97  & 0.71 & 0.62   & 0.90    & 0.69 \\ \hline

    \multirow{5}{*}{\rotatebox{90}{MALLS}}&op-Quanti $\downarrow$     & \textbf{0.94} & 0.99 & 0.95 & 0.95  & 0.99    & 0.96      \\
    &op-negation $\downarrow$   & 0.62 & 0.80 & 0.93 & 0.83  & 0.96   & \textbf{0.09}    \\ 
    &op-AndOr $\downarrow$  & 0.79 & \textbf{0.66} & 0.92 & 0.93  & 0.99        & 0.92  \\
    &t-operator $\downarrow$  & \textbf{0.35} & 0.71 & 0.57 & 0.51  & 0.90        & 0.53 \\
    &t-variable $\downarrow$  & \textbf{0.49} & 0.99  & 0.82 & 0.80   & 0.92    & 0.83 \\\bottomrule
  \end{tabular}}
  \caption{Result when comparing the ground-truth and their perturbations using different metrics. Each row represents results under a specific perturbation (see \S\ref{sec:perturbation}). These numbers indicate the degree of sensitivity of each metric when a specific form of perturbation is applied to the ground-truth. The \textbf{bold} values on each row indicate the metric with the highest sensitivity.}
  \label{tab:perturbation}
  \vspace{-3mm}
\end{table}

\begin{figure}[t]
  \centering
  \begin{subfigure}{\columnwidth}
    \centering
    \includegraphics[width=\linewidth]{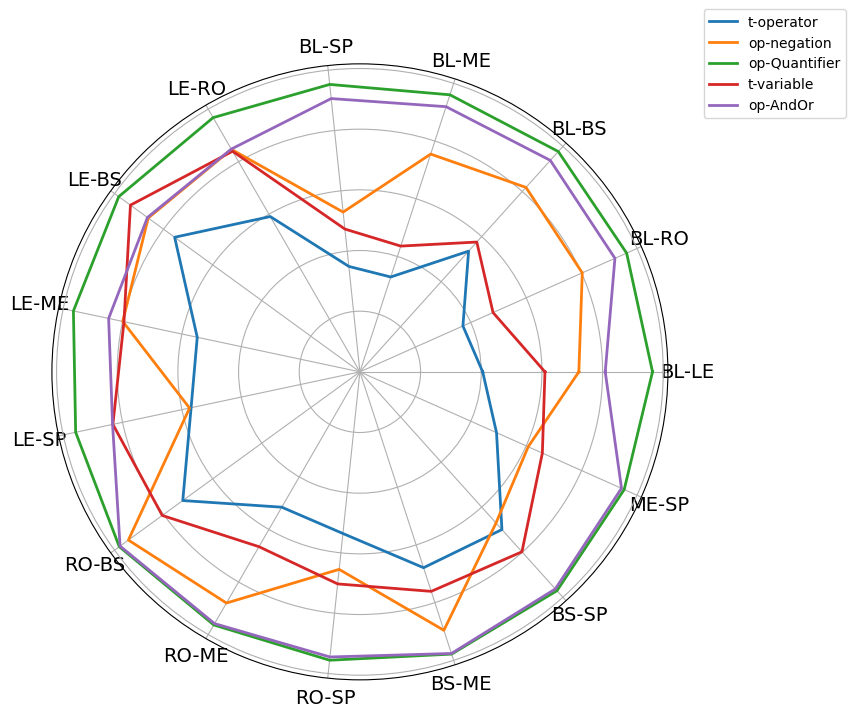}
    \caption*{FOLIO}
  \end{subfigure}
  
  \vspace{5pt} 

  \begin{subfigure}{\columnwidth}
    \centering
    \includegraphics[width=\linewidth]{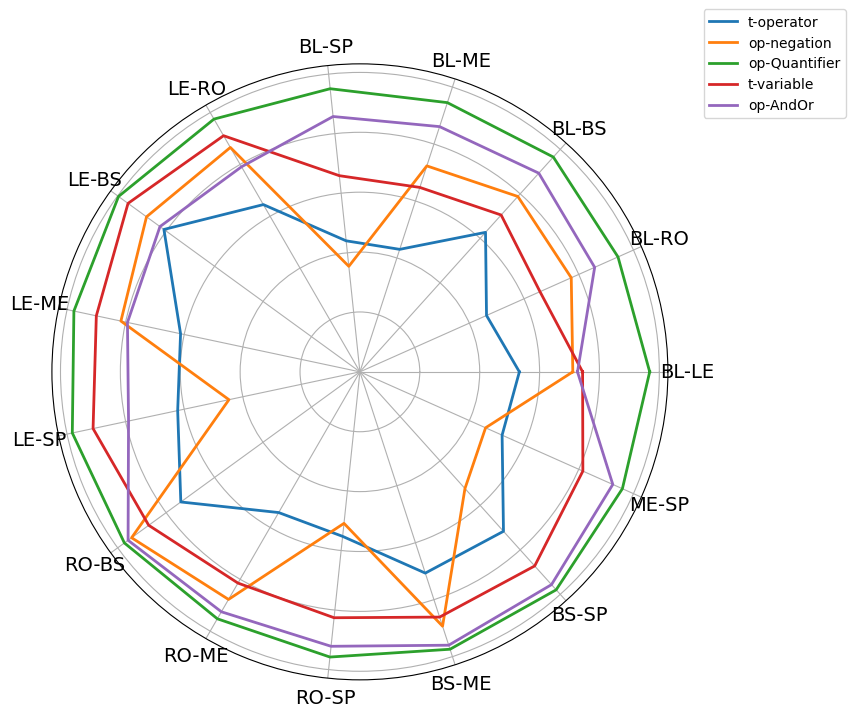}
    \caption*{MALLS}
  \end{subfigure}

  \caption{Extensions of Table~\ref{tab:perturbation} for a combination of 2 metrics. For instance, LE-BS indicates sensitivity of a combined metric interpolating Logical Equivalency~(LE) and BertScore (BS).}
  \label{fig:radial}
  \vspace{-4mm}
\end{figure}

Table~\ref{tab:perturbation} shows the average score by metrics for variations of perturbations. Here, we observe that BLEU score identifies text perturbations more effectively than specialized FOL and graph metrics. LE and Smatch++ exhibit sensitivity to all the perturbations, with a notable sensitivity to the operator-based perturbations. Smatch++ is better suitable to capture the presence or absence of negation operator over LE score, showing a better sensitivity to variations in operators over lexical graph based LE metric. As expected, text-based perturbations should influence translation metric scores, and this is evident in the case of operator and variable perturbations. In contrast, predicate perturbations cause only a minimal drop in scores, as they impact a smaller portion of the dataset, as outlined in Table~\ref{tab:pertperc}. 

In a combination metric setting (Figure~\ref{fig:radial}), Smatch++ paired with textual metrics such as BLEU and METEOR exhibits enhanced sensitivity, particularly when perturbing negation operators—suggesting that changes to operators are more effectively captured through a combination of graph- and text-based metrics. The combination of BLEU and the LE metric shows higher sensitivity to conjunction and disjunction operator. Although BLEU alone is significantly impacted by textual perturbations, combining it with other metrics proves more effective in identifying such textual variations. 
 This trend is also evident for other metric combinations, with scores detailed in Appendix \ref{app:pertpairs}).

\paragraph{Metric Ranking vs. LLM-as-a-judge alignment.}
We now turn to evaluating the alignment of each individual metric\footnote{We perform a qualitative analysis to show disagreements between individual metrics for ranking the samples in Appendix~\ref{app:qual}, where there is a clear distinction between text- and operator-based metrics.}, as well as the combination of two metrics, with the LLM (o3-mini) judge.

As shown in Table~\ref{tab:rankings} (on the diagonal), Bertscore demonstrates a closer alignment with LLM judge. LE score shows the weakest alignment, but improves when combined with BertScore. The results suggest that, despite the low alignment of structured evaluators such as LE score and Smatch++, using other metrics alongside help with improving their alignment, making them suitable for both semantic and syntactic verification.
\begin{table}\footnotesize
  \centering
      \resizebox{1\columnwidth}{!}{%
  \begin{tabular}{lllllllll}
    \toprule
    \multicolumn{2}{c}{RMSE} & {\textbf{BL}} 
    & {\textbf{LE}} 
    & {\textbf{RO}} 
    & {\textbf{ME}} 
    &{\textbf{BS}} 
    & {\textbf{SP}}  \\ 
    \hline
    \multirow{6}{*}{\rotatebox{90}{FOLIO}} &\textbf{BL} & \hl{1.01} & 0.97 & 0.90 & 0.89 & \textbf{0.83} & 0.94  \\
    &\textbf{LE} & - & \hl{1.13} & 0.94 & 0.92 & \textit{0.88} & 0.98  \\
    &\textbf{RO} & - & - & \hl{0.92} & 0.89 & \textit{0.84} & 0.93  \\
    &\textbf{ME} & - & - & - & \hl{0.90} & \textbf{0.83} & 0.92  \\
    &\textbf{BS} & - & - & - & - & \textit{\hl{0.85}} & \textit{0.86}  \\
    &\textbf{SP} & - & - & - & - & - & \hl{0.98}  \\
    \hline
    \multirow{6}{*}{\rotatebox{90}{MALLS}}&\textbf{BL} & \hl{0.85} & 0.86 & 0.84 & 0.83 & \textit{0.78} & 0.86  \\
    &\textbf{LE} & - & \hl{1.03} & 0.87 & 0.87 & \textit{0.79} & 0.92  \\
    &\textbf{RO} & - & - & \hl{0.85} & 0.82 & \textbf{0.77} & 0.84  \\
    &\textbf{ME} & - & - & - & \hl{0.85} & \textbf{0.77} & 0.84  \\
    &\textbf{BS} & - & - & - & - & \textit{\hl{0.79}} & \textit{0.80}  \\
    &\textbf{SP} & - & - & - & - & - & \hl{0.92}  \\
    \bottomrule
  \end{tabular}}
  \caption{{RMSE scores comparing metric-based and LLM-based (o3-mini) rankings over 728 records (1 ground-truth + 3 FOL candidates). Diagonal values (highlighted) show individual metrics vs. LLM rankings; off-diagonal entries show combined metrics vs. LLM. 
  } }
  \label{tab:rankings}
  \vspace{-3mm}
\end{table}

\section{Conclusion}

 This study has explored the sensitivity of various metrics in evaluating the closeness of generated First-Order Logic (FOL) translations of natural language statements and the ground-truth.
 By carefully analyzing the sensitivity of existing metrics through perturbations of ground-truth FOLs, we identified critical gaps in commonly used metrics. Commonly used FOL metrics are not sufficient for handling anomalies in FOL generation while BertScore in isolation and in combination with other metrics offer the most reliable measure. 

\paragraph{Future Work.}
The conclusions presented in this paper highlight the opportunity to choose a sensitivity- or alignment-based metric depending on the application. The identified metric combinations can serve as a reward function to support the development of translation models or formal generation systems in reasoning spaces. We also emphasize that a dependable automatic metric is essential, not only for rigorous evaluation but also for training the next generation of models that accurately translate natural language into formal symbolic representations. This need is particularly urgent for Autoformalization~\cite{wu2022autoformalization}. We hope the systematic evaluation framework presented here will encourage the development of even more reliable metrics.

\section*{Limitations}
We recognize that GPT models used in our experiments are continually evolving, which may lead to variations in results over time. To manage the computational cost of generating multiple samples, we limited the data sample used in the experiments to 3 samples. This work could ideally be extended to a larger dataset or used as a reference for achieving high performance in existing methodologies, but not as a standalone solution. The FOLIO dataset, despite being widely used, may contain errors inherent to human judgement. 

\bibliography{alignmentFOL/custom}

\appendix
\newpage
\section{Data Statistics}\label{app:stats}
\begin{figure}[ht]
  \includegraphics[width=\columnwidth]{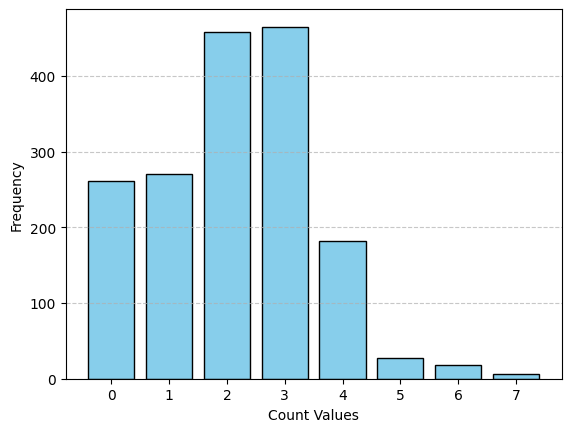}
  \caption{Plot showing the distribution of the operators in each records for FOLIO. The majority of records contain 2-3 operators. Records with 0 operators reflect the presence of a single predicate, indicating no logical connections, while records with 7 operators represent complex statements.}
  \label{fig:dist}
\end{figure}

The FOLIO training set contains 1001 records with ground truth FOLs. 
Upon review, we observe that the number of operators in the records ranges from 0 to 7, with 0 representing a standalone predicate. 
By expanding the data, we observe additional operator combinations for a given sentence. For each set of operators, we generate four sentence variations. The details on the distribution of operators can be referred to in Figure~\ref{fig:dist}. The perturbation effect on FOL records are reported in Table \ref{tab:pertperc}.

\begin{figure}[ht]
  \includegraphics[width=\columnwidth]{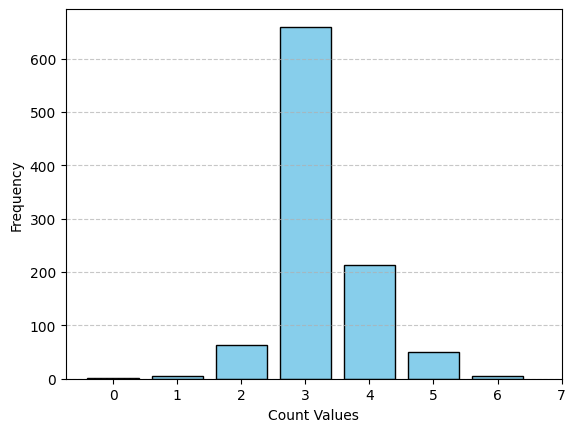}
  \caption{Plot showing the distribution of the operators in each records for MALLS dataset. The majority of records contain 3 operators.}
  \label{fig:dist-malls}
\end{figure}

MALLS training set consists of 27,284 individual FOL statements. For ease of experiment, we randomly sample 1000 statements from the dataset. The distribution of operators is as presented in Figure \ref{fig:dist-malls}.

\section{Rationale for T-Operator}
\label{app:rationale}
The "Text minus Operator" perturbation may not reflect a common surface-level error directly. However, its purpose was to simulate critical logical omissions, which are difficult to isolate using small surface edits.

In LLM generations, we frequently observe mismatches or omissions of logical operators. Table ~\ref{tab:rationale} shows examples from real LLM outputs where the operator in the generated FOL deviates from the ground truth. These reflect structural inconsistencies similar to our perturbation and help justify its inclusion in the sensitivity analysis. 

\begin{table*}[ht]
\centering
\resizebox{\textwidth}{!}{%
\renewcommand{\arraystretch}{1.5}
\begin{tabular}{p{2.0cm}p{4.2cm}p{4.8cm}p{4.8cm}}
\toprule
\textbf{ID} & \textbf{NL Premise} & \textbf{Ground Truth (FOL)} & \textbf{LLM Generation (FOL)} \\
\midrule
FOLIO\_t\_56 & The L-2021 monitor is either used in the library or has a type-C port & Monitor(l\_2021) $\land$ (In(l\_2021, library) $\oplus$ Have(l\_2021, typeCPort)) & Used(L\_2021, Library) $\lor$ HasTypeC(L\_2021) \\
FOLIO\_t\_254 & Two major types of reasoning rules are inductive and deductive reasoning & $\forall$ x (MajorArgumentForm(x) $\rightarrow$ (InductiveReasoning(x) $\oplus$ DeductiveReasoning(x))) & InductiveReasoning(x) $\lor$ DeductiveReasoning(x) \\
\bottomrule
\end{tabular}}
\caption{A sample of Ground Truth vs. LLM Generated FOL Expressions.}
\label{tab:rationale}
\end{table*}

\section{Sample Generation}\label{app:prompt}
We generate the FOL samples from a given NL using the prompt 
\begin{quote}
``\emph{{Given a natural language sentence, your task is to convert the sentence into first-order logic statements using the following operators:} $\land, \lor, \neg, \rightarrow, \leftrightarrow, \forall, \exists, =, \oplus$. {The output is a single first-order statement representing the sentence with no additional tasks.}}''
\end{quote}
where gpt4o provides 3 samples in the form presented in Figure~\ref{fig:sample}.

\begin{figure}[ht]
    \centering
    \fbox{
        \parbox{0.9\linewidth}{
        Given a natural language sentence, your task is to convert the sentence into first-order logic statements using the following operators: 
        \[
        \land, \lor, \neg, \rightarrow, \leftrightarrow, \forall, \exists, =, \oplus.
        \]
        The output is a single first-order statement representing the sentence with no additional tasks. Generate 3 different samples of output.\\
        \\
        \textbf{Text:} All eels are fish.\\
        \textbf{Output:} \hl{1. $\forall x (\text{Eel}(x) \rightarrow \text{Fish}(x))$ 2. $\forall x (\text{E}(x) \rightarrow \text{F}(x))$ 3. $\forall x (\text{IsEel}(x) \rightarrow \text{IsFish}(x))$}
        }
    }
    \caption{Example of sample generation using \texttt{gpt-4o}. The highlighted text is the output from the LLM.}
      \label{fig:sample}
\end{figure}

\section{Human Annotation Details}\label{sec:annotator}
The 87 samples were ranked by human annotators on a scale of [1, 2, 3], where 1 indicates the best match and 3 the least match to the ground truth. Additionally, we conducted an LLM-based ranking in which the LLM is prompted to rank the same three samples using the same scale. To determine alignment between LLM and human judgments, we compute the RMSE between the LLM-predicted rankings and human annotations. Based on the RMSE scores shown in Table~\ref{tab:model_scores}, we select \texttt{o3-mini} as the evaluation LLM due to its closest alignment with human preferences.

\begin{table}[ht]
\centering
\begin{tabular}{l|c}
\hline
\textbf{Model} & \textbf{Score} \\
\hline
gpt-4o         & 0.87 \\
o3-mini        & 0.65 \\
gemini-flash-2 & 0.89 \\
\hline
\end{tabular}
\caption{Human Alignment with existing LLMs, where the lowest RMSE score represents better alignment}
\label{tab:model_scores}
\end{table}

\paragraph{Annotators qualifications}
We enlisted three annotators with at least a Master's degree in CS or AI to rank the similarity between the ground truth FOL and the generated samples. The instructions provided to the experts were kept open-ended, offering only a basic overview of the logic and ranking criteria to avoid inducing bias. Although the instructions suggested ranking randomly in case of a tie, we deduplicated the values and assigned the same rank to the matching FOLs, as described in the previous passage. 
\paragraph{Pairwise Ranking.}
A pairwise comparison is performed between the three human annotations to determine the final rankings. For each pair of annotations, we compare their relative rankings to establish the overall order. This approach ensures that the final ranking is derived by consistently evaluating each annotation against the others in a pairwise manner. 
To do this, we calculate the scores for each FOL and compute the average score for each sentence. These averages are then processed to obtain the final value.
\paragraph{Annotator Instruction}
{The surface level similarity is not specifically a restriction in scoring as we intended to avoid imposing strict biases and to allow for a diverse evaluation. Below is the instruction in detail.}

{The task is to rank the first-order logic (FOL) translations for a given ‘gold label’ a rank of [1,2,3], where 1 represents the best match and 3 represents a comparatively bad match to the gold FOL. You are given 3 variations of FOL for each sentence. Please feel free to rank based on your preference.} {Few good-to-know instructions:}
\begin{itemize}
[noitemsep,leftmargin=4mm,parsep=0pt,partopsep=0pt]
    \item $F_1 \land F_2$: Logical AND, True only if both $F_1$ and $F_2$ are true
    \item $F_1 \lor F_2$: Logical OR, False if both $F_1$ and $F_2$ are false
    \item $\neg$: Negation
    \item $\rightarrow$: Implies
    \item $\Leftrightarrow$ Double Implies
    \item $\forall$: For All quantifier
    \item $\exists$: There Exists quantifier
    \item $=$: Equals 
    \item $F_1 \oplus F_2$: XOR, True only if $F_1$ or $F_2$ are true
\end{itemize}

{If two FOLs are the ‘same’, randomly number them.} \textit{Ex:} $F_1$: $A \land B$ Rank 3, $F_2$: $A \land B$ Rank 2, $F_3$: $A \rightarrow B$ Rank 1.

{You can lower the rank for structure (syntax) or grammar (semantic) errors. Please do not change the format of the file. Just add the rank next to ‘Rank’ for each FOL.}

There are one or more correct rankings. In case of ‘all incorrect’, pick the rank based on the closest match to the gold FOL.

{Example (put your ranking at the end of each statement after “Rank”):}
\begin{itemize}
[noitemsep,leftmargin=4mm,parsep=0pt,partopsep=0pt]
    \item     label: $\forall$ x ({Square}(x) $\rightarrow$ {Shape}(x))
    \item     FOL1: $\forall$ x ({Square}(x) $\rightarrow$ {Shape}(x)) Rank: 1
    \item     FOL2: $\forall$x ($\neg${Shape}(x) $\rightarrow$ $\neg${Square}(x)) Rank: 2
    \item     FOL3: $\forall$ x ({Squares}(x) $\rightarrow$ {Shapes}(x)) Rank: 3
\end{itemize}

A few input examples along with real-time human annotations are presented in Table~\ref{tab:fol_samples}.

\begin{table*}[ht]
    \centering
    \small
          \resizebox{1\textwidth}{!}{%
    \renewcommand{\arraystretch}{1.5}
    \begin{tabular}{>{\raggedright\arraybackslash}p{11cm}|ccc}
        \hline
        Samples & H1 & H2 & H3 \\
        \hline
        \textbf{label:} $\forall$ x ($\neg$ WantToBeAddictedTo (x, caffeine) $\rightarrow$ $\neg$ AwareThatDrug(x, caffeine))& [3,2,1] & [3,2,1] & [3,2,1] \\
        \textbf{fol1:} $\forall$ x ($\neg$ Wants(x, Addicted(Caffeine)) $\rightarrow$ Knows(x, Drug(Caffeine))) \\
        \textbf{fol2:} $\forall$ x ($\neg$ Wants(x, Addicted(Caffeine)) $\rightarrow \neg$ Unaware(x, Drug(Caffeine))) \\
        \textbf{fol3:} $\forall$ x ($\neg$ Wants(x, Addicted(Caffeine)) $\rightarrow$ Aware(x, Drug(Caffeine))) \\
        
        \hline
        \textbf{label:} $\forall x (Eel(x) \rightarrow Fish(x))$& [3,2,1] & [3,2,1] & [3,2,1] \\
        \textbf{fol1:} $\forall x (E(x) \rightarrow F(x))$\\
        \textbf{fol2:} $\forall x (IsEel(x) \rightarrow IsFish(x))$\\ 
        \textbf{fol3:} $\forall x (Eel(x) \rightarrow Fish(x))$ \\
        
        \hline
        \textbf{label:} $\forall x (Outside(x, solarSystem) \oplus In(x, solarSystem))$  & [1,3,2] & [1,3,2] & [1,2,3] \\
        \textbf{fol1:} $\forall x (OutsideSolarSystem(x) \lor InSolarSystem(x))$\\
        \textbf{fol2:} $\forall x ((\neg InSolarSystem(x)) \lor InSolarSystem(x))$\\ 
        \textbf{fol3:} $\forall x (\neg(OutsideSolarSystem(x) \land InSolarSystem(x)))$  \\
        \hline
    \end{tabular}}
    \caption{Three Samples of 3 Human annotators rankings of 3 FOLs. }
    \label{tab:fol_samples}
\end{table*}

\section{LLM Judge Ranking Prompt}\label{app:ranking}
 The used prompt is presented in Figure~\ref{fig:rank}.

\begin{figure}[ht]
    \centering
    \fbox{
        \parbox{0.9\linewidth}{
        Given a ground truth first-order logic statement and three variations of samples, your task is to rank the samples in order of similarity with the label. The output should be a single list with 3 integers, including [1, 2, 3], where 1 represents the closest match and 3 is the least match. Do not include any other explanation and the output form is [rank\_sample1, rank\_sample2, rank\_sample3].\\
        \\
        \textbf{Label:} $\forall x (\text{Eel}(x) \rightarrow \text{Fish}(x))$\\
        \textbf{Sample 1:} $\forall x (\text{E}(x) \rightarrow \text{F}(x))$\\
        \textbf{Sample 2:} $\forall x (\text{IsEel}(x) \rightarrow \text{IsFish}(x))$\\
        \textbf{Sample 3:} $\forall x (\text{Eel}(x) \rightarrow \text{Fish}(x))$\\
        \textbf{Output:} \hl{[1, 3, 2]}
        }
    }
  \caption{Example of the prompt used for ranking the FOLs using the LLM judge. The \hl{highlighted} text is the output from the LLM.}
  \label{fig:rank}
\end{figure}

\section{Pairwise Perturbations}
\label{app:pertpairs}
To study the effect of perturbation on the combinations, we obtain sensitivity scores as shown across Table \ref{tab:quant} to Table \ref{tab:variable}. When compared to a single metric, the combination helps with improving the sensitivity of the metric.
\begin{table}\footnotesize
  \centering
      \resizebox{1\columnwidth}{!}{%
  \begin{tabular}{lllllllll}
    \toprule
    \multicolumn{2}{c}{op-Quantifier} & {\textbf{BL}}
    & {\textbf{LE}} 
    & {\textbf{RO}} 
    & {\textbf{ME}} 
    &{\textbf{BS}} 
    & {\textbf{SP}}  \\ 
    \hline
    \multirow{6}{*}{\rotatebox{90}{FOLIO}} 
    &\textbf{BLEU}       & \hl{0.96} & 0.96 & 0.96 & 0.96 & 0.98 & 0.95 \\
    &\textbf{LE}         & -         & \hl{0.97} & 0.97 & 0.97 & 0.98 & 0.96 \\
    &\textbf{Rouge}      & -         & -    & \hl{0.97} & 0.96 & 0.98 & 0.96 \\
    &\textbf{Meteor}     & -         & -    & -    & \hl{0.96} & 0.98 & 0.95 \\
    &\textbf{BertScore}  & -         & -    & -    & -    & \hl{1.00} & 0.97 \\
    &\textbf{Smatch++}   & -         & -    & -    & -    & -    & \hl{0.95} \\
    \hline
    \multirow{6}{*}{\rotatebox{90}{MALLS}} 
    &\textbf{BLEU}       & \hl{0.94} & 0.97 & 0.94 & 0.95 & 0.97 & 0.95 \\
    &\textbf{LE}         & -         & \hl{1.00} & 0.97 & 0.98 & 1.00 & 0.98 \\
    &\textbf{Rouge}      & -         & -    & \hl{0.95} & 0.95 & 0.97 & 0.96 \\
    &\textbf{Meteor}     & -         & -    & -    & \hl{0.95} & 0.97 & 0.96 \\
    &\textbf{BertScore}  & -         & -    & -    & -    & \hl{0.99} & 0.98 \\
    &\textbf{Smatch++}   & -         & -    & -    & -    & -    & \hl{0.97} \\
    \hline
  \end{tabular}}
  \caption{Quantifier perturbation pairwise comparison}
  \vspace{-10pt}
  \label{tab:quant}
\end{table}

\begin{table}\footnotesize
  \centering
      \resizebox{1\columnwidth}{!}{%
  \begin{tabular}{lllllllll}
    \toprule
    \multicolumn{2}{c}{op-Negation} & {\textbf{BL}}
    & {\textbf{LE}} 
    & {\textbf{RO}} 
    & {\textbf{ME}} 
    &{\textbf{BS}} 
    & {\textbf{SP}}  \\ 
    \hline
    \multirow{6}{*}{\rotatebox{90}{FOLIO}} 
    &\textbf{BLEU}       & \hl{0.68} & 0.72 & 0.80 & 0.75 & 0.82 & 0.53 \\
    &\textbf{LE}         & -         & \hl{0.77} & 0.85 & 0.80 & 0.86 & 0.57 \\
    &\textbf{Rouge}      & -         & -    & \hl{0.93} & 0.88 & 0.94 & 0.65 \\
    &\textbf{Meteor}     & -         & -    & -    & \hl{0.83} & 0.90 & 0.61 \\
    &\textbf{BertScore}  & -         & -    & -    & -    & \hl{0.96} & 0.67 \\
    &\textbf{Smatch++}   & -         & -    & -    & -    & -    & \hl{0.38} \\
    \hline
    \multirow{6}{*}{\rotatebox{90}{MALLS}} 
    &\textbf{BLEU}       & \hl{0.62} & 0.71 & 0.77 & 0.72 & 0.79 & 0.35 \\
    &\textbf{LE}         & -         & \hl{0.80} & 0.87 & 0.82 & 0.88 & 0.45 \\
    &\textbf{Rouge}      & -         & -    & \hl{0.93} & 0.88 & 0.94 & 0.51 \\
    &\textbf{Meteor}     & -         & -    & -    & \hl{0.83} & 0.89 & 0.46 \\
    &\textbf{BertScore}  & -         & -    & -    & -    & \hl{0.96} & 0.52 \\
    &\textbf{Smatch++}   & -         & -    & -    & -    & -    & \hl{0.09} \\
    \hline
  \end{tabular}}
  \caption{Negation perturbation pairwise comparison}
  \vspace{-10pt}
  \label{tab:negation}
\end{table}

\begin{table}\footnotesize
  \centering
      \resizebox{1\columnwidth}{!}{%
  \begin{tabular}{lllllllll}
    \toprule
    \multicolumn{2}{c}{op-AndOr} & {\textbf{BL}}
    & {\textbf{LE}} 
    & {\textbf{RO}} 
    & {\textbf{ME}} 
    &{\textbf{BS}} 
    & {\textbf{SP}}  \\ 
    \hline
    \multirow{6}{*}{\rotatebox{90}{FOLIO}} 
    &\textbf{BLEU}       & \hl{0.88} & 0.81 & 0.92 & 0.92 & 0.94 & 0.91 \\
    &\textbf{LE}         & -         & \hl{0.74} & 0.85 & 0.85 & 0.87 & 0.83 \\
    &\textbf{Rouge}      & -         & -    & \hl{0.96} & 0.96 & 0.98 & 0.95 \\
    &\textbf{Meteor}     & -         & -    & -    & \hl{0.96} & 0.98 & 0.94 \\
    &\textbf{BertScore}  & -         & -    & -    & -    & \hl{1.00} & 0.96 \\
    &\textbf{Smatch++}   & -         & -    & -    & -    & -    & \hl{0.93} \\
    \hline
    \multirow{6}{*}{\rotatebox{90}{MALLS}} 
    &\textbf{BLEU}       & \hl{0.79} & 0.73 & 0.86 & 0.86 & 0.89 & 0.86 \\
    &\textbf{LE}         & -         & \hl{0.66} & 0.79 & 0.79 & 0.83 & 0.79 \\
    &\textbf{Rouge}      & -         & -    & \hl{0.92} & 0.93 & 0.96 & 0.92 \\
    &\textbf{Meteor}     & -         & -    & -    & \hl{0.93} & 0.96 & 0.92 \\
    &\textbf{BertScore}  & -         & -    & -    & -    & \hl{0.99} & 0.96 \\
    &\textbf{Smatch++}   & -         & -    & -    & -    & -    & \hl{0.92} \\
    \hline
  \end{tabular}}
  \caption{And-Or perturbation pairwise comparison}
  \vspace{-10pt}
  \label{tab:andor}
\end{table}

\begin{table}\footnotesize
  \centering
      \resizebox{1\columnwidth}{!}{%
  \begin{tabular}{lllllllll}
    \toprule
    \multicolumn{2}{c}{t-Operator} & {\textbf{BL}}
    & {\textbf{LE}} 
    & {\textbf{RO}} 
    & {\textbf{ME}} 
    &{\textbf{BS}} 
    & {\textbf{SP}}  \\ 
    \hline
    \multirow{6}{*}{\rotatebox{90}{FOLIO}} 
    &\textbf{BLEU}       & \hl{0.19} & 0.40 & 0.37 & 0.33 & 0.54 & 0.35 \\
    &\textbf{LE}         & -         & \hl{0.62} & 0.59 & 0.55 & 0.75 & 0.57 \\
    &\textbf{Rouge}      & -         & -    & \hl{0.56} & 0.52 & 0.72 & 0.54 \\
    &\textbf{Meteor}     & -         & -    & -    & \hl{0.47} & 0.68 & 0.49 \\
    &\textbf{BertScore}  & -         & -    & -    & -    & \hl{0.89} & 0.70 \\
    &\textbf{Smatch++}   & -         & -    & -    & -    & -    & \hl{0.51} \\
    \hline
    \multirow{6}{*}{\rotatebox{90}{MALLS}} 
    &\textbf{BLEU}       & \hl{0.35} & 0.53 & 0.46 & 0.43 & 0.63 & 0.44 \\
    &\textbf{LE}         & -         & \hl{0.71} & 0.65 & 0.61 & 0.81 & 0.62 \\
    &\textbf{Rouge}      & -         & -    & \hl{0.58} & 0.54 & 0.74 & 0.55 \\
    &\textbf{Meteor}     & -         & -    & -    & \hl{0.51} & 0.71 & 0.52 \\
    &\textbf{BertScore}  & -         & -    & -    & -    & \hl{0.90} & 0.72 \\
    &\textbf{Smatch++}   & -         & -    & -    & -    & -    & \hl{0.53} \\
    \hline
  \end{tabular}}
  \caption{Operator perturbation pairwise comparison}
  \vspace{-10pt}
  \label{tab:operator}
\end{table}

\begin{table}\footnotesize
  \centering
      \resizebox{1\columnwidth}{!}{%
  \begin{tabular}{lllllllll}
    \toprule
    \multicolumn{2}{c}{t-variable} & {\textbf{BL}}
    & {\textbf{LE}} 
    & {\textbf{RO}} 
    & {\textbf{ME}} 
    &{\textbf{BS}} 
    & {\textbf{SP}}  \\ 
    \hline
    \multirow{6}{*}{\rotatebox{90}{FOLIO}} 
    &\textbf{BLEU}       & \hl{0.25} & 0.61 & 0.48 & 0.44 & 0.58 & 0.47 \\
    &\textbf{LE}         & -         & \hl{0.97} & 0.84 & 0.80 & 0.94 & 0.83 \\
    &\textbf{Rouge}      & -         & -    & \hl{0.71} & 0.67 & 0.81 & 0.70 \\
    &\textbf{Meteor}     & -         & -    & -    & \hl{0.62} & 0.76 & 0.66 \\
    &\textbf{BertScore}  & -         & -    & -    & -    & \hl{0.90} & 0.80 \\
    &\textbf{Smatch++}   & -         & -    & -    & -    & -    & \hl{0.70} \\
    \hline
    \multirow{6}{*}{\rotatebox{90}{MALLS}} 
    &\textbf{BLEU}       & \hl{0.49} & 0.74 & 0.66 & 0.65 & 0.70 & 0.66 \\
    &\textbf{LE}         & -         & \hl{1.00} & 0.91 & 0.90 & 0.96 & 0.91 \\
    &\textbf{Rouge}      & -         & -    & \hl{0.83} & 0.81 & 0.87 & 0.83 \\
    &\textbf{Meteor}     & -         & -    & -    & \hl{0.80} & 0.86 & 0.82 \\
    &\textbf{BertScore}  & -         & -    & -    & -    & \hl{0.92} & 0.87 \\
    &\textbf{Smatch++}   & -         & -    & -    & -    & -    & \hl{0.83} \\
    \hline
  \end{tabular}}
  \caption{Variable perturbation pairwise comparison}
  \vspace{-10pt}
  \label{tab:variable}
\end{table}

\section{Qualitative Study}
\label{app:qual}
While identifying cases where all metrics strongly disagree at the same time is unlikely, we highlight two examples where pairwise disagreements are notable. These disagreements are calculated by taking an average of the absolute difference between the score of the samples. A higher value indicates larger disagreement.

$$\text{Disagreement} = \frac{1}{N} \sum_{i=1}^{N} \left| s_i^{(1)} - s_i^{(2)} \right|$$

Where N is the number of samples, $s_i^{(X)}$ is the score assigned to the i-th sample of a metric X.

In the table Table~\ref{tab:qualitative}, we picked two examples.
FOLIO\_t\_461 shows divergence between text-based metrics, while BERTScore(BS) assigns partial credit for variations such as “WorkFullTime”  vs “Work,” BL would fail to capture this similarity.
FOLIO\_t\_1098 contrasts semantic and structural metrics, where logical metrics like SP and LE give high scores, while text-based metrics disagree due to semantic differences.

\begin{table*}[ht]
\resizebox{1\textwidth}{!}{%
\centering
\small
\renewcommand{\arraystretch}{1.2}
\begin{tabular}{p{1.8cm}|p{0.9cm}|p{0.6cm}|p{3.0cm}|p{3.4cm}|p{3.4cm}}
\hline
\textbf{ID} & \textbf{Metric} & \textbf{Score} & \textbf{NL Premise} & \textbf{Ground Truth (FOL)} & \textbf{LLM Generation (FOL)} \\
\hline
FOLIO\_t\_461 & BL-BS & 0.94 & Those who are enrolled in an academic program can not work full-time & 
$ \forall$ x (EnrolledIn(x, academicProgram) $\rightarrow$ $\lnot$ Work(x, fullTime)) & 
1. $\forall$ x (EnrolledInAcademicProgram(x) $\rightarrow$ $\lnot$ WorkFullTime(x)) \\
& & & & & 2. $\forall$ x (EnrolledInAcademicProgram(x) $\rightarrow$ $\lnot$ CanWorkFullTime(x)) \\
& & & & & 3. $\forall$ x (Enrolled(x) $\rightarrow$ $\lnot$ FullTimeWork(x)) \\
\hline
FOLIO\_t\_1098 & BL-SP & 0.76 & If people don't care about cleanliness, then they do not prioritize cleaning & 
$\forall$ x ($\lnot$ CareAbout(x, cleanliness) $\rightarrow$ $\lnot$ Prioritize(x,\ cleaning)) & 
1. $\forall$ x\ ($\lnot$ CaresAboutCleanliness(x) $\rightarrow$ $\lnot$ PrioritizesCleaning(x)) \\
& & & & & 2. $\forall$ x ($\lnot$ CareAboutCleanliness(x) $\rightarrow$ $\lnot$ PrioritizeCleaning(x)) \\
& & & & & 3. $\forall$ x ($\lnot$ C(x) $\rightarrow$ $\lnot$ P(x)) \\
\hline
\end{tabular}}
\caption{Qualitative analysis of disagreement between metrics for samples generated by the LLM. Score indicates the disagreement score calculated for that metric combination.}
\label{tab:qualitative}
\end{table*}

\section{Package Usage}
This paper utilizes automatic evaluation metrics and datasets in compliance with their respective licenses. Specifically, we employ BLEU, BertScore (MIT License), ROUGE (Apache-2.0 License), METEOR (MIT License), Logical Equivalence (Apache-2.0 License), and Smatch++ (GNU General Public License). The datasets FOLIO and MALLS, used in this research, are open-sourced under the MIT License and CC-BY-NC-4.0 License respectively.

The packages used in this paper are primarily sourced from the evaluation metrics provided by \href{https://github.com/huggingface/evaluate/tree/main/metrics}{Hugging Face's Evaluate library}. Additionally, the source code for Logical Equivalence and Smatch++ was utilized.

\end{document}